\documentclass[sigconf]{acmart}
\AtBeginDocument{%
  }

\usepackage{balance} 
\usepackage{graphicx}
\usepackage{subcaption}
\usepackage{mdframed}
\usepackage{adjustbox}

\setcopyright{none}
\renewcommand\footnotetextcopyrightpermission[1]{}
\copyrightyear{2026}
\acmYear{2026}
\acmDOI{}
\acmConference[AIDA2J]{Artificial Intelligence for Access to Justice, Dispute Resolution, and Data Access}{June 08,
  2026}{Singapore}
\acmISBN{none}

\begin{document}

\title[Investigating Multi-Agent Deliberation in Law]{Investigating Multi-Agent Deliberation in Law}

\author{Cor Steging}
\email{c.c.steging@rug.nl}
\orcid{0000-0001-6887-1687}
\affiliation{%
  \institution{Bernoulli Institute of Mathematics, Computer Science and Artificial Intelligence, University of Groningen}
  \country{The Netherlands}
}

\author{Ludi van Leeuwen}
\email{l.s.van.leeuwen@rug.nl}
\orcid{0000-0003-3165-4376}
\affiliation{%
  \institution{Bernoulli Institute of Mathematics, Computer Science and Artificial Intelligence, University of Groningen}
  \country{The Netherlands}
}

\author{Tadeusz Zbiegień}
\email{tadeusz.zbiegien@doctoral.uj.edu.pl}
\orcid{0000-0001-9052-6978}
\affiliation{%
  \institution{Department of Legal Theory, Jagiellonian University}
  \country{Poland}
}


\renewcommand{\shortauthors}{Steging et al.}


\begin{abstract}
Artificial Intelligence is increasingly applied to the field of law, and has the potential to increase access to justice. One particular movement that is gaining traction is that of agentic AI, wherein AI agents, based on Large Language Models (LLMs) can take autonomous actions. In particular, multi-agent approaches in the legal domain remain largely unexplored. 
In this paper, we investigate multi-agent deliberation methods for legal reasoning tasks using LLMs. 
We explore multi-agent deliberation (MAD) and introduce two novel multi-agent frameworks inspired by courtroom procedures and legal argumentation.
Our experiments on both legal and non-legal benchmarks reveal that multi-agent frameworks achieve comparable overall performance to baseline large language models, but produce significantly distinct answers. 
Notably, these approaches can successfully solve cases that the baseline fails to address, and vice versa.
We conduct a qualitative evaluation and highlight scenarios where multi-agent frameworks outperform monolithic approaches.
For example, multi-agent approaches appear better suited for answering questions that require critical thinking from multiple perspectives.
Our work positions multi-agent systems as a promising direction for AI in the legal domain, while demonstrating the potential of law-inspired multi-agent approaches for deliberation. 
\end{abstract}




\maketitle

\section{Introduction}

Artificial Intelligence (AI), and in particular Large Language Models (LLMs), have seen increasing adoption in the legal domain, for example, to annotate legal texts~\cite{Savelka2023TheUnreasonable},
enhancing the usability of court forms~\cite{steenhuis2023beyond}
and assisting professionals by streamlining legal procedures~\cite{LLLM_survey1, Lai2024Large}. 
These AI developments have the potential to improve access to justice by assisting both laypersons and legal professionals.
A particularly important area within the legal domain is legal reasoning, a complex and multifaceted process that demands a diverse set of cognitive and analytical skills~\cite{Ellsworth2005LegalReasoning}.

At the same time legal reasoning differs fundamentally from tasks such as drafting or improving usability. Instead, it involves interpretation, weighing values, and navigating open-textured concepts~\cite{bench-capon_model_2003, bench-capon_oward_nodate, bench-capon_computational_nodate, hart_concept_2012, verheij_formalizing_2016}. In real legal practice, opposing parties advance arguments that both support their own position and attempt to undermine those of the other side, with a third neutral party making the final decision. As a result, legal reasoning inherently involves multiple perspectives that must be weighed and reconciled. This deliberative and adversarial nature suggests that computational models for legal reasoning could benefit from adopting a multi-agent approach, where distinct viewpoints are explicitly represented and evaluated.

However, current approaches based on LLM to legal deliberation~\cite{goebel2024overview}, typically rely on a single model that produces a single narrative. This one-sided reasoning risks tunnel vision, overlooking alternative interpretations, and leading to less robust conclusions. In contrast, multi-agent frameworks can capture and evaluate competing perspectives more explicitly. In a standard Multi-Agent Deliberation (MAD) setup~\cite{du2023improving}, for instance, several agents independently address a question and then reflect on each other’s responses, enabling richer and more balanced reasoning. This comes at a time when agentic AI approaches are being adopted increasingly across both the legal domain and other fields~\cite{acharya2025agentic}, yet the potential of multi-agent approaches remains largely unexplored in legal research.

In this paper, we therefore explore multi-agent approaches to legal deliberation. 
We evaluate standard MAD frameworks on these tasks and also introduce two novel multi-agent frameworks based on legal procedures and argumentation. 
The first is a 3-Ply framework, inspired by the three-ply approach in~\cite{rissland_case-based_1987, ashley_modelling_1988}, in which agents adopt the roles of plaintiff and defendant in a courtroom, with a third agent acting as judge to weigh the merits of each side. This design explicitly explores two competing perspectives and is procedurally suggestive for dispute-resolution settings, since it explicitly models adversarial party positions and a neutral adjudicative role
The second, the Parrots framework, stages a dialogue between a primary agent and a set of critical `parrot' agents, each representing distinct perspectives grounded in argumentation theory~\cite{ParrotPaper}. 
Together with a baseline LLM classifier, we evaluate these frameworks across five benchmarks, consisting of four legal reasoning tasks and one logical reasoning task. 
Our goal is to examine the potential of multi-agent approaches for legal reasoning and to introduce two novel, legal-inspired frameworks for multi-agent deliberation. 


\begin{figure*}[ht]
    \centering
    \begin{subfigure}[b]{0.19\textwidth}
        \includegraphics[width=\textwidth]{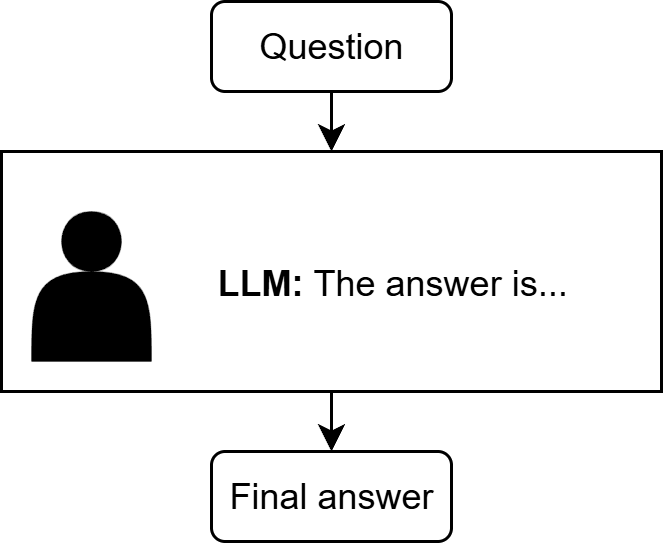}\vspace{2pt}
        \caption{Baseline}
        \label{fig:image1}
    \end{subfigure}
    \hfill
    \begin{subfigure}[b]{0.23\textwidth}
        \includegraphics[width=\textwidth]{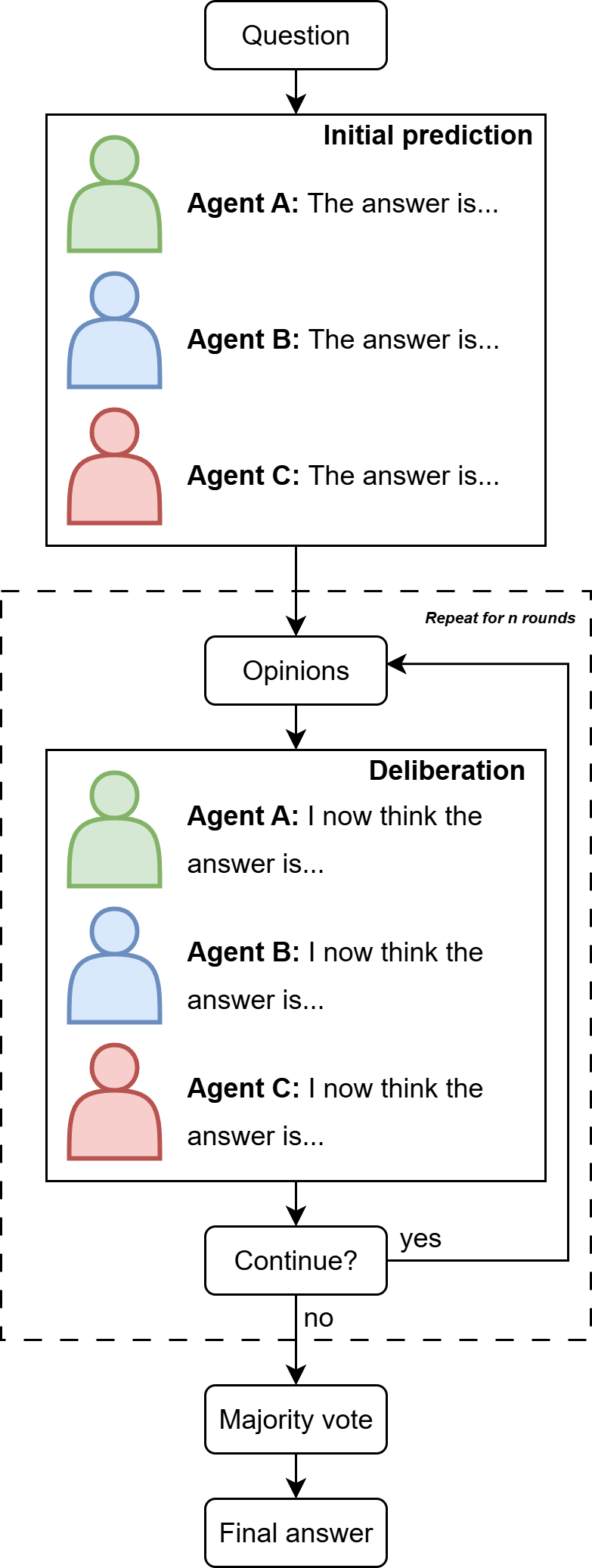}\vspace{2pt}
        \caption{Multi-Agent Deliberation}
        \label{fig:image2}
    \end{subfigure}
    \hfill
    \begin{subfigure}[b]{0.19\textwidth}
        \includegraphics[width=\textwidth]{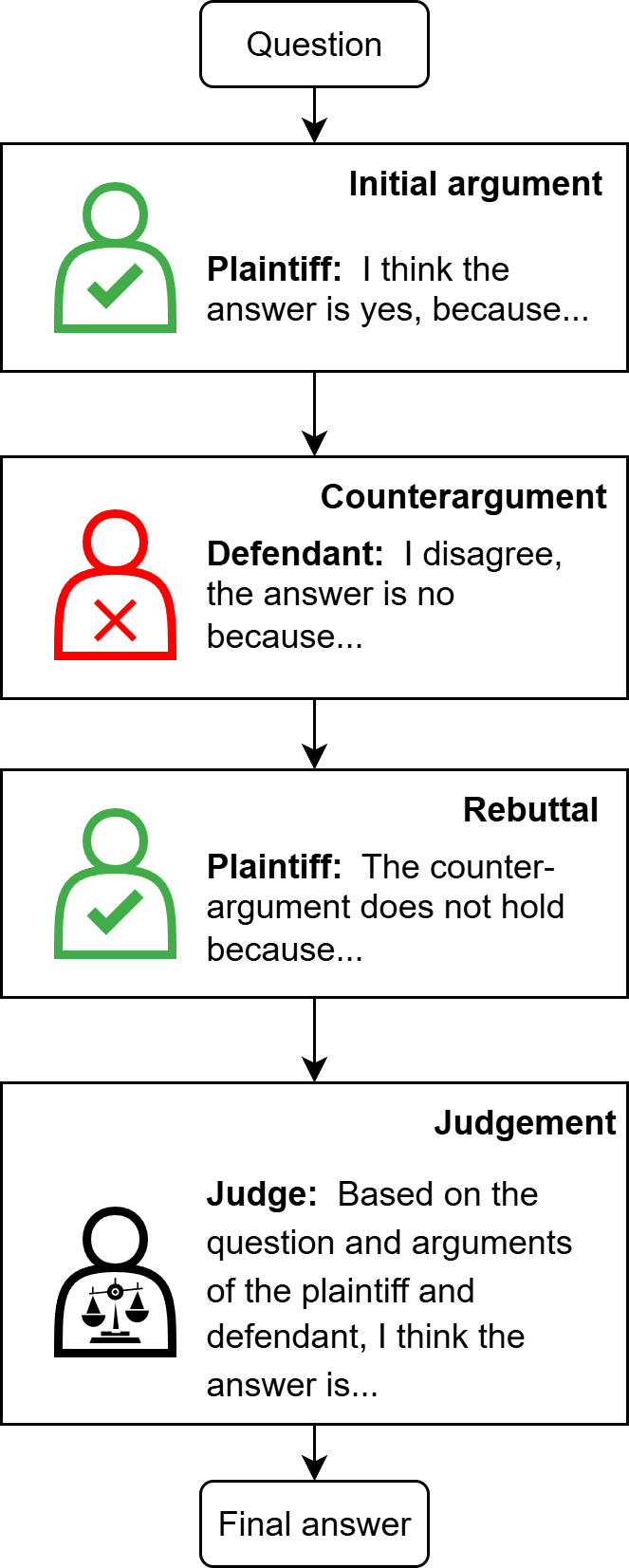}\vspace{2pt}
        \caption{3-Ply}
        \label{fig:image3}
    \end{subfigure}
    \hfill
    \begin{subfigure}[b]{0.23\textwidth}
        \includegraphics[width=\textwidth]{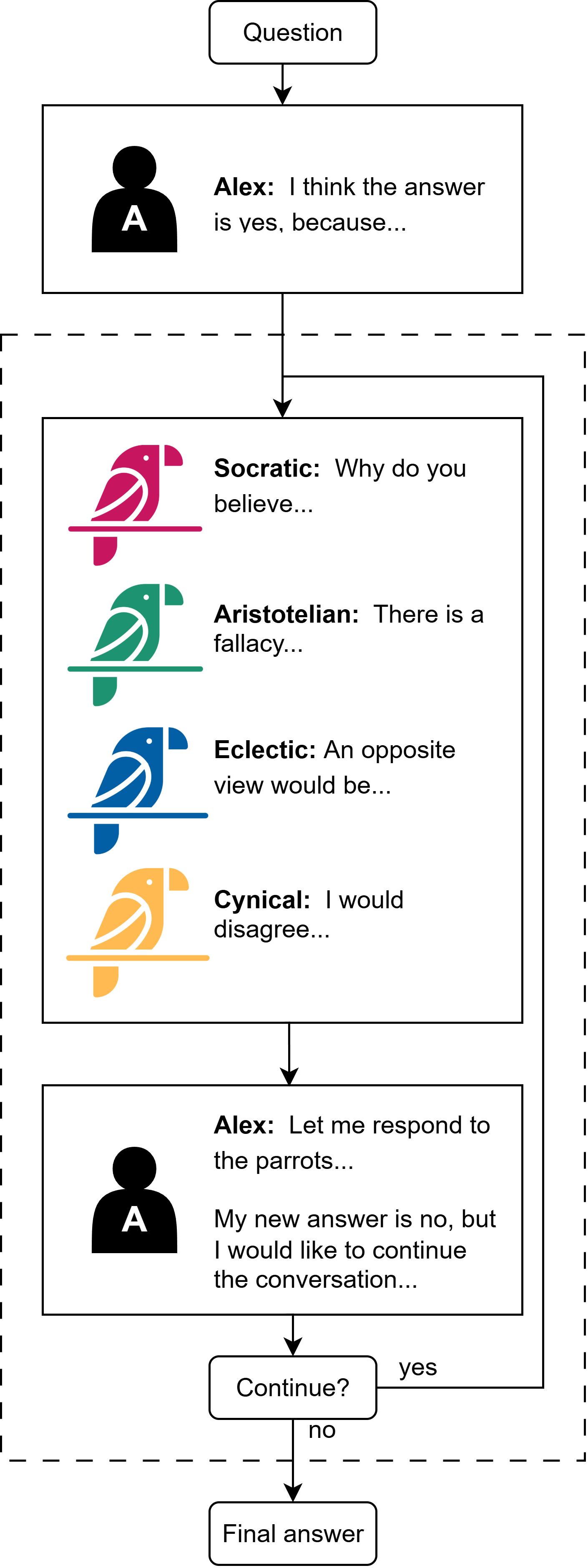}\vspace{2pt}
        \caption{Parrots}
        \label{fig:image4}
    \end{subfigure}
    \caption{The pipelines of each of the frameworks we evaluate.}
    \label{fig:pipelines}
\end{figure*}

\section{Background}
While there has been a recent surge in the commercial adoption of AI within the legal domain, the academic subdiscipline of AI \& Law has been active for over 30 years \cite{Governatori2022ThirtyYears, Sartor2022ThirtyYears, Villata2022ThirtyYears}. Early work focused on formal models of legal reasoning and argumentation, gradually shifting towards a focus on machine learning and large language models in recent years \cite{Governatori2022ThirtyYears, Sartor2022ThirtyYears, Villata2022ThirtyYears, Lai2024Large}. 


The potential of multi-agent based approaches for studying legal phenomena has gained gradual attention in recent years, though it remains a relatively niche area \cite{Benthall2021Agent, Schwartz2020Agent}.
Previous studies largely concentrated on simulating procedures and arguments, whereas our approach emphasizes deliberative reasoning.
\citeauthor{Arisaka2022Multi} (\citeyear{Arisaka2022Multi}) propose an approach to formalize legal reasoning using multi-agent abstract argumentation. 
\citeauthor{zhang2025mitigatingmanipulationenhancingpersuasion} (\citeyear{zhang2025mitigatingmanipulationenhancingpersuasion}) developed a multi-agent method for legal argument generation that produces 3-Ply arguments, demonstrating the advantages of multi-agent architectures over baseline models.
Similarly, \citeauthor{Jiang2025AgentsBench} (\citeyear{Jiang2025AgentsBench}) implemented a judicial bench deliberation process to mirror real-world legal dynamics and evaluate multi-agent architectures in legal judgment prediction tasks. 
Notably, \citeauthor{Jiang2025AgentsBench} applied different types of agents to represent roles on a judicial bench, distinguishing between professional and lay judges. 
Although these approaches aim to simulate the procedural and argumentative dynamics of real-world legal reasoning, our work instead focuses on applying deliberation-based architectures to solve concrete legal reasoning tasks.
Additional role-inspired and agentic approaches in the legal domain have been explored in both recent works \cite{he2024agentscourtbuildingjudicialdecisionmaking, jiang2024agentsbenchlargelanguage, devadiga2025samvadmultiagentsimulatingjudicial} and earlier investigations \cite{10.5555/1671082.1671098, Schwartz2019Agent}.
Beyond these works, legal question-answering systems have until recently predominantly relied on non-agentic methods, including standard machine learning, language models, retrieval-based architectures, or hybrid approaches combined with legal ontologies \cite{StegingICAIL2025, Abdallah_2023, Martinez-Gil2023Survey, goebel2024overview}.

Our approach builds on well-established literature in multi-agent systems, as well as scholarship highlighting the inherently argumentative nature of law. A long line of jurisprudence and AI \& Law research emphasizes that legal conclusions are justified through structured, deliberative argumentation~\cite{RotoloSartor2023ArgumentationExplanationLaw, PrakkenSartor2015LawLogic, Feteris1997SurveyLegalArgumentation}. Research on formal argumentation and deliberative processes within multi-agent systems offers a useful framework to study or simulate the argumentative dynamics of legal processes \cite{Arisaka2022Multi, Benthall2021Agent, Schwartz2020Agent, MaudetParsonsRahwan2007}. 
We build on these insights to develop an architecture where cooperating and competing agents exchange structured arguments for legal question answering. From an access-to-justice and dispute resolution perspective, exposing competing positions and their justifications can better support informed participation. 

\section{Method}
In our experiment, we compare three multi-agent frameworks to a baselines large language model: a standard \textit{Multi-Agent-Deliberation (MAD)} framework, a \textit{3-Ply} framework based on court room procedures, and a \textit{Parrot} framework, based on ideas from the field of argumentation. 
We explore the different approaches by comparing their performance and answers on a set of five datasets, consisting of four legal reasoning tasks and one logical reasoning task. 
The latter was chosen to demonstrate the novel legal-inspired approaches are domain-agnostic and can thus also solve non-legal tasks.
All tasks are binary in nature and require the model to answer either `yes' or `no'.
In this section, we describe the frameworks and the datasets. A high-level overview of the frameworks can be seen in Figure~\ref{fig:pipelines}. The full prompts for each system and how they are employed can be found in the appendix.

\subsection{Deliberation Frameworks}
\paragraph{Baseline} The Baseline classifier is a simple framework used to compare with the three multi-agent frameworks. In this setup, a question is given to a large language model that should give an answer and a short explanation. This Baseline classifier acts as a reference point to better grasp the performance and behaviors of the multi-agent frameworks. Questions are answered using a single model call.

\paragraph{Multi-Agent Deliberation (MAD)} Our Multi-Agent Deliberation (MAD) classifier employs three large language models, referred to as agents, each responsible for producing a prediction. 
The framework is grounded in prior work inspired by the Society of Mind~\cite{du2023improving}. Over a fixed number of deliberation rounds, each agent is shown the responses of the other two and asked to revise its own accordingly. In our implementation, the agents make an initial prediction and then deliberate for two rounds. The final answer is then determined by majority vote. Given three initial predictions and two rounds of three refinements each, the MAD setup requires a total of nine model calls.

\paragraph{3-Ply} We introduce the 3-Ply classifier, based on a simulated courtroom with three LLM agents representing the plaintiff, defendant, and judge. The plaintiff aims to convince the judge that the answer to the question is `yes', whereas the defendant will claim that the answer is `no'. The impartial judge will then decide, based on the merits of the arguments of the plaintiff and defendant, what the answer to the question should be and why. 
The reasoning behind this 3-Ply approach is that both perspectives are always considered explicitly. 
The 3-Ply classifier process consists of three parts: an initial argument by the plaintiff, a counterargument by the defendant, and finally a rebuttal by the plaintiff. The judge uses the arguments put forth across all parts to determine the final answer. 
This leads to a total of four model calls to answer questions using the 3-Ply framework. 

\paragraph{Parrots} The Parrot classifier is a novel framework that we introduce, based on a recent notion in the field of argumentation that large language models should argue with us by design~\cite{ParrotPaper}. Instead of an agreeable stochastic parrot, proponents of this notion suggest a set of four critical parrots, each with a different perspective. We apply this notion to multi-agent deliberation, where an artificial agent named `Alex', represented by an LLM, argues back and forth with a set of four parrots, represented by a single LLM. Following the literature, parrots challenge assumptions or definitions used by Alex or other parrots (Socratic), rebut or undermine the arguments of Alex and other parrots’, testing their robustness (Cynical), offer alternative perspectives or possible interpretations that might otherwise be overlooked (Eclectic), and critically evaluate the logical strength of the arguments, highlighting weaknesses, inconsistencies or fallacies (Aristotelian). In our framework, the Alex agent makes an initial prediction in the same fashion as the Baseline classifier. The parrots react to this prediction, each from their own perspective. Subsequently, Alex responds to the parrots and is asked whether they want to continue the conversation or not. In our experiment, Alex may continue the conversation for a maximum of three rounds. The number of model calls per question is, therefore, varied between three and seven.

\subsection{Datasets}
We evaluate the four frameworks discussed across five datasets designed to benchmark legal and logical reasoning capabilities. Each dataset is described in detail below.

\begin{figure}
    \begin{mdframed}
        \footnotesize
        Classify if the description of each clause is correct. \\
        
        \textbf{Clause:} Geolocation data, which is any Information which indicates the geographical position of your device and which may include the time the location was recorded, and may include precise (e.g., GPS-level) location data. Some Apalon Apps may collect this data when the App is not open or actively in use (This may help us or third parties to customize information, content, or offers to you, for instance.) The Apalon App and/or your mobile device may provide options for you to control the use and/or sharing of precise device location data.\\
        
        \textbf{Description:} The policy describes collection of the user's WiFi location data by a party to the contract.\\
        
        Is the description of the clause correct (yes) or not (no)?\\
        Reason and end your answer with "Answer: yes or no"
        \end{mdframed}
    \caption{Example question 869 from the Privacy dataset.}
\label{fig:example_question}
\end{figure}

\paragraph{LEXAM} This dataset was built from law school examinations across a wide range of courses and jurisdictions and aims to test legal reasoning~\cite{fan2025lexambenchmarkinglegalreasoning}. We limit ourselves to the multiple-choice subset, which contains 1,660 questions with four answer options each. Questions are in English or German and pertain to Swiss or international law, dealing with criminal, public, private, or interdisciplinary law. Since the current implementation of our frameworks can only deal with binary tasks, we restructure the multiple-choice format into binary `yes'/`no' questions. 

\paragraph{COLIEE} This benchmark consists of bar exam questions that evaluate legal entailment in the context of Japanese Civil Law~\cite{goebel2024overview}. Questions consist of legal articles and a statement, and the task is to determine whether the articles entail the statement. Questions have been translated into English by the owners of the dataset (2025 version).

\paragraph{SARA} This dataset is designed to test reasoning about U.S. federal tax law~\cite{holzenberger2021factoring}. It provides statutes together with cases. Each case is paired with an entailment prompt that asks whether a specific statutory section applies to the facts. The SARA dataset is also part of the LegalBench benchmark collection for legal reasoning \cite{legalbench}.

\paragraph{PRIVACY} This dataset contains questions about privacy policies~\cite{zimmeck2019maps}. The questions consist of a clause from a privacy policy along with a candidate description of that clause (for example, a statement that the policy covers the collection of cookies, flash cookies, pixel tags, or comparable identifiers by a contractual party). The model must decide whether the description accurately reflects the clause or not. The Privacy dataset is also contained in the LegalBench collection~\cite{legalbench}.

\paragraph{LOGIQA} This is a multiple-choice reading comprehension dataset containing logical reasoning questions from the Chinese National Civil Servants Examination, designed to test critical thinking and problem-solving~\cite{10174688}. Each instance provides a short passage, a query, and four candidate answers, with the task of selecting the correct one. As with LEXam, we convert the questions into a binary format. We use the English version translated from the original exam materials.

\begin{figure*}[t]
    \centering
    \includegraphics[width=1\linewidth]{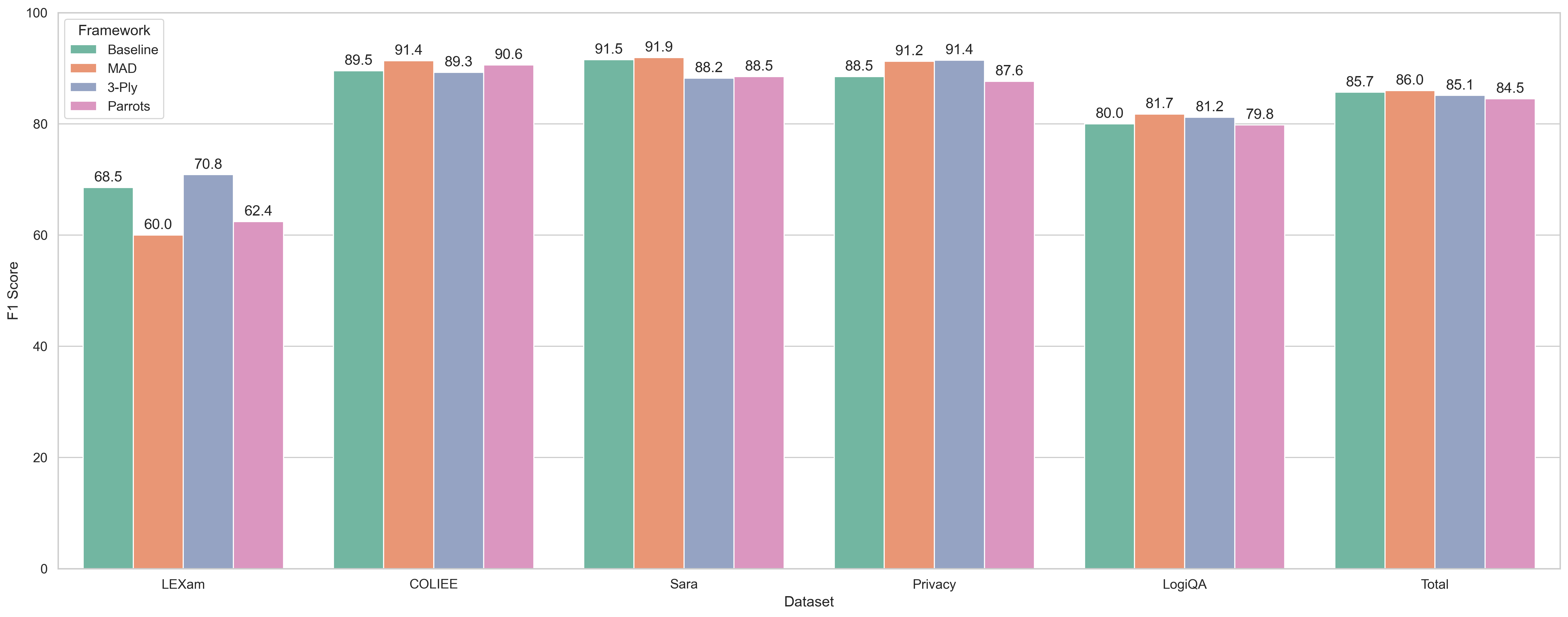}
    \caption{Performance comparison of the multi-agent frameworks against baseline across the five benchmark datasets.}
    \label{fig:main_comparison}
\end{figure*}

We perform no further pre-processing on the datasets. 

\subsection{Experimental setup}
In our experiment, we randomly sample 250 questions from each of the datasets, yielding a total of 1250 questions for evaluating our deliberation frameworks. We ensure a balanced label distribution in these questions. The four frameworks are tasked with solving these questions. To illustrate, we show an example question from the Privacy dataset in Figure~\ref{fig:example_question}. Here we see a clause from a privacy policy, alongside a description. The question asks whether that description is correct with regard to the clause. In this example question, the answer is `yes' because under this particular privacy policy geolocation data includes WiFi location data. Hence, even though the WiFi location data is not directly mentioned, it is deemed to be included under the broader clause, which illustrates typical interpretative reasoning in which lawyers engage. 

To improve performance, we make use of 2-shot prompting, where we provide the frameworks with two similar examples from a separate training dataset in addition to the main question. We use BM25 ranking to identify the two most similar questions, a standard information retrieval method to estimate document–query relevance~\cite{Amati2009}.

We used OpenAI's GPT-5-mini model, as their models are among the most used. For financial reasons, we did not explore larger and more expensive models, which we address in the discussion. The prompts used by the frameworks were created based on the literature wherever possible and empirical results based on light prompt engineering using a smaller model (gpt-5-nano). For instance, the system prompt that describes the perspectives of the parrots agent was taken verbatim from the paper that introduced the concept~\cite{ParrotPaper}.


\section{Results}
In Figure~\ref{fig:main_comparison}, we show the F1 scores of the baseline and three multi-agent framework on each of the five datasets, as well as the aggregated total performance across all datasets.

\begin{figure}[t]
    \centering
    \includegraphics[width=0.85\linewidth]{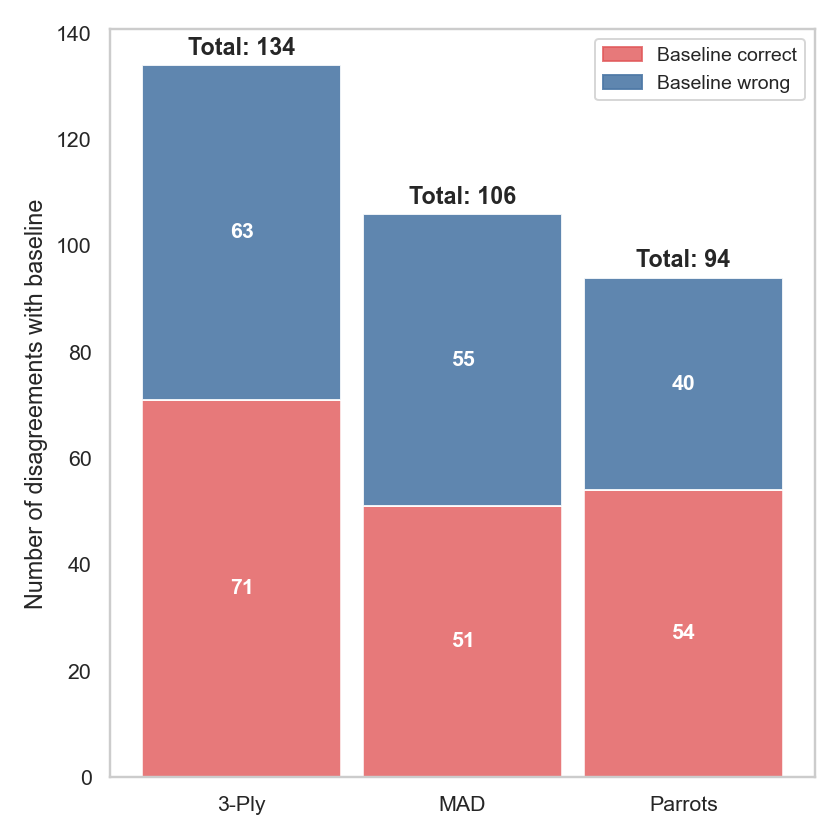}
    \caption{Number of questions in which the framework disagreed with the baseline.}
    \label{fig:disagreements}
\end{figure}

Our analysis shows that all three frameworks differ significantly from the baseline in their decision patterns (McNemar test, p < 0.001). These differences do not result in significant performance gains, however, as evident in Figure~\ref{fig:main_comparison}.
However, the frameworks are able to correctly resolve a subset of cases that the baseline fails to solve (3-Ply: 47.0\%, MAD: 51.9\%, Parrots: 42.6\% of disagreements), as shown in Figure~\ref{fig:disagreements}.

We illustrate the effects of the number of agents and reflection rounds in the MAD framework in Table~\ref{tbl:mad_comparison}, where we plot the performance of the framework on the Privacy dataset, using different parameters.

Furthermore, we show the effect of the number of deliberation rounds in the Parrot framework in Table~\ref{tbl:parrot_comparison}. In this setup, we let Alex and the Parrots either argue for a static number of rounds (1-3), or let Alex decide when to end the conversation (dynamic).

\begin{figure*}[htbp]
\centering

\hfill
\begin{minipage}[t]{0.49\linewidth}
    \centering
    \begin{subfigure}[t]{\linewidth}
        \begin{mdframed}
            \footnotesize
            The clause refers to geolocation/GPS-level device location and control over precise location sharing; it does not mention Wi‑Fi based location data or explicitly state Wi‑Fi as a source. \\ \\
            \textbf{Final answer: no} 
        \end{mdframed}
        \caption{Baseline response.}
        \label{fig:baseline_response}
    \end{subfigure}
    
    \vspace{2.1em}

    \begin{subfigure}[t]{\linewidth}
        \begin{mdframed}
            \footnotesize
\textbf{Plaintiff (Initial argument):}

1) The clause expressly covers "Geolocation data" defined as any information indicating the geographical position of your device — a definition that is method-agnostic and therefore includes location derived from WiFi access points.\\
2) The parenthetical "may include precise (e.g., GPS-level) location data" uses "e.g.", signaling that GPS is an example, not an exhaustive list; other common positioning methods such as WiFi-based location fall within the broad category. \\
3) Because WiFi-derived coordinates are a recognized form of device geolocation, asserting that the policy describes collection of the user's WiFi location data is a correct and reasonable interpretation of the clause.\\
\textbf{Answer: yes}\\

\textbf{Defendant (Counterargument):}

No. Critique:
1) The clause does not explicitly mention WiFi-derived location (...)\\
2) Broad definitions are ambiguous (...)\\
3) The parenthetical example (“may include precise (e.g., GPS-level) location data”) weakens, not strengthens, the plaintiff’s claim (...)\\
4) The description adds unstated specificity about the collector relationship (...)\\
\textbf{Answer: no}\\

\textbf{Plaintiff (Rebuttal):}

I respond to the defendant’s points and defend that the description is correct: the clause’s broad definition of “Geolocation data” reasonably and textually encompasses WiFi-derived location data collected by a contracting party (...)
1) On “no explicit mention of WiFi” — (...) That language is method‑agnostic and unqualified. A correct description need not reproduce every possible technical method when the policy uses an intentionally broad, inclusive definition.\\
2) On ambiguity of broad definitions — the task is whether the description is correct, not whether the clause lists every technology. (...)\\
3) On the parenthetical example (“may include precise (e.g., GPS‑level)”) — the use of “e.g.” and “may include” signals an illustrative, non‑exhaustive list. (...)\\
4) On the collector relationship. The description’s formulation (“by a party to the contract”) is a fair characterization of that explicit language.\\
(...)
Conclusion: because the clause’s broad, unqualified definition of geolocation data, its illustrative parenthetical, and its explicit statement that Apalon apps may collect such data together provide clear textual support, it is correct to describe the policy as covering collection of the user’s WiFi location data by a party to the contract.

\textbf{Answer: yes}\\

\textbf{Judge (Judgement):} \\
(...) The clause uses an intentionally broad, method‑agnostic definition: "any Information which indicates the geographical position." That wording is not limited to particular technologies and therefore logically encompasses location inferred from WiFi access points. (...). The task is whether the description is correct, not whether the clause lists every technical mechanism.\\
Plaintiff’s points (broad, method‑agnostic definition; non‑exhaustive example; explicit statement that Apalon may collect geolocation) are logically sound and directly support the description.\\
Defendant’s points highlight the lack of explicit wording for WiFi but do not overcome the clear inclusive definition in the clause.\\The clause’s broad definition of geolocation plus the explicit statement that Apalon apps may collect that data provides textual support that WiFi-derived location data would be included and that a party to the contract may collect it. 
\\
\\
\textbf{Final answer: yes}

        \end{mdframed}
        \caption{3-Ply response.}
        \label{fig:three_ply_response}
    \end{subfigure}
\end{minipage}
\hfill
\begin{minipage}[t]{0.49\linewidth}
    \centering
    \begin{subfigure}[t]{\linewidth}
        \begin{mdframed}
            \footnotesize
            \textbf{Agent A voted `yes', because:}\\
            The clause explicitly states Wattpad "may also collect the account holder's cell phone number if the account holder opts in to an SMS alert service," which is a direct statement that the service (a party to the contract) may collect the user's phone number. \\
            
            \textbf{Agent B voted `no', because:}\\
            The clause states that no personal information is collected, while the description asserts the policy collects the user's phone number, which is clearly personal information. These two statements conflict, so the description is not correct.\\
            
            \textbf{Agent C voted `yes', because:}\\
            Reason: The clause covers "geolocation data" — any information indicating the geographical position of your device — and explicitly gives GPS as an example of a precise method. Wi‑Fi–based location is another common method of deriving device geolocation, so collecting Wi‑Fi location falls within the clause's broad definition of geolocation data even though Wi‑Fi is not named explicitly.\\
            
            \textbf{Final answer: yes}
                    \end{mdframed}
                    \caption{MAD response.}
                    \label{fig:mad_response}
                \end{subfigure}
    \vspace{0.00005em}

    \begin{subfigure}[t]{\linewidth}
        \begin{mdframed}
            \footnotesize
            \textbf{Alex}: No. The clause does not specifically mention WiFi-based location (...) “WiFi location” is a common method of deriving device geolocation and therefore falls under the general term “geolocation data.”
            
            \textbf{Answer: yes}\\
            
 \textbf{Socratic Parrot}: You assume "geolocation data" necessarily includes WiFi-derived locations. Is the task's annotation guideline allowing inference from a broad category (geolocation) to specific methods (WiFi), or does it require explicit mention of the method? \\

\textbf{Aristotelian Parrot}: Logically, the clause explicitly names "geographical position" and "precise (e.g., GPS-level) location data." That supports inclusion of methods that indicate position, but the argument is inductive — it depends on whether the labeler’s rule treats methods as implicit members of the category. Point out this dependence before asserting "yes." \\

\textbf{Cynical Parrot}: Be cautious: if the annotation scheme treats only explicitly named data types as matches, calling it "WiFi location" could be marked wrong. Alex's appeal to common practice (WiFi is a common method) may not satisfy a strict labeler. \\

\textbf{Eclectic Parrot}: Practical compromise: if the dataset uses semantic grouping (any location method counts), answer yes; if it requires explicit mention, answer no. Alex should state which convention he's using and justify that choice. \\

 \textbf{Alex}:\\
 Response to Socratic : The example annotations provided show that a broad category ("geolocation data") is treated as covering specific technical methods. Do you want me to cite those examples as the basis for this interpretive rule? \\
 Response to Aristotelian: Agreed — if the labelling rule is logical grouping rather than literal mention, the inference is valid; the dataset examples indicate that's the applied rule. \\
 Response to Cynical: Point taken — under a strictly literal scheme this would be wrong. Do you want me to flag this as a potential ambiguous case for reviewers? \\
Response to Eclectic: I accept the practical compromise: apply semantic grouping. Therefore WiFi-derived location is covered by “geolocation data.” \\

\textbf{Alex's refined argument}: (...) The clause covers any information indicating device location, not limited to a single technology. Example annotations treat geolocation broadly, mapping it to specific methods. Therefore WiFi-derived location, being a common method to determine device position, is reasonably included.\\

\textbf{Final Answer: Yes}
        \end{mdframed}
        \caption{Parrot response.}
        \label{fig:parrot_response}
    \end{subfigure}
\end{minipage}

\caption{Answers to example question (Figure~\ref{fig:example_question}).}
\label{fig:example_responses}
\end{figure*}

\begin{table}[h!]
\centering
\caption{Performance of the MAD framework with varying numbers of agents and deliberation rounds on the Privacy dataset, and (correct) disagreement with baseline.}
\label{tbl:mad_comparison}
\begin{tabular}{rrccc}
\toprule
\textbf{\#Agents} & \textbf{\#Rounds} & \textbf{F1 Score} & \textbf{Disagreement} & \textbf{Correct} \\
\midrule
3  & 0 & 89.43 & 8  & 6  \\
10 & 0 & 88.99 & 7  & 5  \\
2  & 1 & 90.30 & 10 & 8  \\
3  & 1 & 89.42 & 10 & 7  \\
3  & 2 & 90.31 & 15 & 11 \\
\bottomrule
\end{tabular}
\end{table}

\begin{table}[h!]
\centering
\caption{Performance of the Parrots framework with static and dynamic deliberation rounds on the Privacy dataset, and (correct) disagreement with baseline.}
\label{tbl:parrot_comparison}
\begin{tabular}{rccc}
\toprule
\textbf{\# Rounds} & \textbf{F1 Score}  & \textbf{Disagreement} & \textbf{Correct} \\
\midrule
1       & 87.66 & 12 & 6 \\
2       & 88.55 & 10 & 6 \\
3       & 87.22 & 9  & 4 \\
Dynamic & 86.74 & 16 & 7 \\
\bottomrule
\end{tabular}
\end{table}

We qualitatively evaluated the outputs of the four frameworks on a random sample of questions, examining not only their final answers but also the accompanying justifications. To illustrate this, Figure~\ref{fig:example_responses} presents the responses of all four frameworks to example question 869 from the Privacy dataset (Figure~\ref{fig:example_question}).

\section{Discussion}
Based on the performance of the four frameworks, as shown in Figure~\ref{fig:main_comparison}, no single framework consistently outperforms the others. In two of the datasets, the 3-Ply framework achieves the highest F1 score, while the MAD framework performs best on the remaining three. Across all datasets, the average F1 scores are highly similar, differing by no more than 1.5 points. 

The absence of a significant performance difference does not imply that the frameworks made similar predictions. A significant difference was observed between the predictions of each multi-agent framework and those of the baseline classifier. As illustrated in Figure~\ref{fig:disagreements}, approximately 7.5–10\% of the questions received different answers from the multi-agent frameworks compared to the baseline. In roughly half of these cases, the baseline’s answer was incorrect, whereas the multi-agent framework provided the correct response. 
Furthermore, we observe an interesting asymmetry in the Privacy dataset: many questions that the baseline fails to solve are correctly resolved by at least one of the multi-agent frameworks, whereas the reverse does not occur; every case correctly solved by the baseline is also solved by at least one multi-agent framework. Although this comparison reflects the combined behavior of multiple frameworks rather than a one-to-one evaluation, it suggests that multi-agent approaches can capture forms of reasoning that remain inaccessible to the baseline classifier, even though they also introduce their own errors.

We illustrate this phenomenon in Figure~\ref{fig:example_responses}, which presents the outputs of the four frameworks for the example question shown in Figure~\ref{fig:example_question}.
In this example, the Baseline produces an incorrect answer by failing to focus on the fact that the clause does not explicitly mention WiFi location data.
This highlights a recurring pattern in the results, where frameworks need to balance between literal and pragmatic interpretations.
Although MAD ultimately provides the correct answer, it first needs to reconcile conflicting agent responses that include hallucinatory justifications unrelated to the question (for instance, references to cell phone number data).
By contrast, both 3-Ply and Parrots clearly articulate the central issue, favouring a pragmatic interpretation over the literal one adopted by the Baseline. 
Because of the structured nature of their deliberation, these two multi-agent frameworks arrive at a more nuanced, legally informed, and ultimately correct stance.

Our general qualitative observations suggest that multi-agent frameworks are effective in cases where legal clauses are ambiguous or allow multiple interpretations, such as the literal and pragmatic interpretation in Figure~\ref{fig:example_question}. 
In such scenarios, MAD and 3-Ply often achieve higher accuracy than the baseline by reconciling diverse perspectives, while Parrots can identify task-specific patterns in the data that the baseline overlooks. 
Additionally, multi-agent outputs provide richer justifications, which can make hybrid human-AI collaboration more effective.
Hallucinations remain a key limitation in all frameworks, as justifications sometimes include entities or conclusions not present in the source text. 
Recent work argues that these hallucinations are structurally inevitable, arising from the fundamental design of large language models~\cite{LLMsWillAlwaysHallucinate}.
We should note that these findings are exploratory, and they highlight the need for standardized evaluation methods for multi-interpretation reasoning in legal AI as a direction for future work.

One of the major downsides of multi-agent approaches to LLMs is that they require more calls and are thus more expensive in terms of resources. While the baseline only requires a single call, 3-Ply uses 4 per question, and MAD uses 9. The Parrot framework used an average of 3.48 calls per question, as it can decide to end the deliberation dynamically. In terms of pure performance, the extra calls used by the multi-agent approaches do not seem to improve the reasoning of the models.

In Table~\ref{tbl:mad_comparison}, we more closely examined the MAD framework using different numbers of agents and rounds of reflection. 
We see that performance remains rather stable across the different parameters. For instance, a setup with only 3 agents and no reflection rounds yields an F1 score of 89.4, only 0.9 lower than those with two reflection rounds. Note that in the setups without reflection rounds, the MAD classifier effectively acts as an ensemble with majority voting. This suggests that the reflection itself therefore does contribute significantly to the performance. More reflections do seem to increase the number of disagreements with the baseline classifier, including the number of correct disagreements. So while the performance does not seem to change, the reasoning behavior does change slightly. 

We evaluate the effect of the number of deliberation rounds in the Parrots framework in Table~\ref{tbl:parrot_comparison}. Similar to the MAD framework, changing the number of rounds or making it dynamic, does not seem to impact performance significantly. The number of disagreements appears to decrease slightly with more static rounds, and appears higher with a dynamic number of deliberation rounds.

Comparing results across datasets shows that some benchmarks are inherently easier than others. However, with respect to the frameworks, there is no clear distinction between performance on legal versus non-legal benchmarks, as the observed differences are comparable.
A closer analysis of errors reveals that most mistakes across all frameworks were false negatives, with rates of 82.02\% for the Baseline, 82.18\% for MAD, 65.59\% for 3-Ply, and 85.42\% for Parrots. The 3-Ply framework is the exception, showing a notably lower false-negative rate, which may reflect its inherently more polarized setup.
 
We should note that our frameworks were tested using a smaller commercially available language model due to financial constraints. Larger models, with larger context windows, could yield different reasoning behaviour as it can effectively work with more dialogue and information. The behavior of LLM is non-deterministic, and therefore, results may be influenced by noise. 
Furthermore, since all of benchmarks are publicly available on the internet, there is chance of data contamination, and that the LLM may have been exposed to the questions during training, which could yield inflated results. 
We hypothesize that such contamination may have a larger positive impact on monolithic systems than on multi-agent approaches, as the latter benefit from deliberation between agents; however, this remains an open question for future research.

One advantage of a multi-agent approach is that inter-agent reasoning is made explicit through argumentation. However, these arguments are not assessed for soundness or validity, and LLMs are known to produce illogical conclusions even from seemingly coherent arguments~\cite{StegingICAIL2025}. In particular, there is no guarantee that agents act rationally with respect to previously stated arguments, whether produced by other agents or by themselves. At the same time, AI \& Law offers a well-established body of work on formal argumentation. Future research could therefore investigate neuro-symbolic approaches that integrate these formal argumentation methods into multi-agent systems.

Another potential advantage of a multi-agent approach is that it may help mitigate sycophancy, i.e., the tendency of a single model to uncritically agree with the user or with earlier outputs. This effect can arise when agents are encouraged to adopt differing perspectives, as in the MAD approach, and more explicitly in systems such as 3-Ply and Parrots, where agents are assigned specialized roles with predefined viewpoints. Future work should empirically assess under which design choices such role- and perspective-based agent architectures meaningfully reduce sycophantic behavior, and when they merely reproduce it at the group level.





    

\section{Conclusion}
In this study, we explore the potential of multi-agent deliberation in legal reasoning tasks. We find that the multi-agent approaches do not provide a higher overall performance. By providing various perspectives, however, the multi-agent frameworks can solve questions that monolithic approaches cannot, while failing on other. Additionally, we introduce two new multi-agent deliberation frameworks inspired by courtroom procedures and argumentation theory, showing that explicitly incorporating diverse viewpoints can lead to alternative conclusions. 

\begin{acks}
This research was funded by the Hybrid Intelligence Center, a 10-year programme funded by the Dutch Ministry of Education, Culture and Science through the Netherlands Organisation for Scientific Research, https://hybrid-intelligence-centre.nl.  
The research leading to these results has also received funding from the National Science Centre, Poland, project no. 2025/57/N/HS5/01561, titled “Uncertainty and Argumentation: Decision-Making Under Uncertainty in Legal Disputes."
\end{acks}

\bibliographystyle{ACM-Reference-Format}
\bibliography{sample-base}

\clearpage
\appendix
\section*{Appendix}

\section{Prompts}
This section details the prompts used in our three systems. Figure~\ref{fig:MAD_prompts} shows the prompts for the multi-agent deliberation (MAD) system, based on ~\cite{du2023improving}. Each agent first receives the initial prompt in Figure~\ref{fig:MAD_prompts_1}. In later rounds, agents receive the prompt in Figure~\ref{fig:MAD_prompts_2}, which includes their previous response together with the responses of the other agents.

Figure~\ref{fig:3ply_prompts} presents the prompts used in the 3-Ply system. Figures~\ref{fig:3ply_judge_SP}, \ref{fig:3ply_plaintiff_SP}, and~\ref{fig:3ply_defendant_SP} show the system prompts for the judge, plaintiff, and defendant agents. Figures~\ref{fig:3ply_promts_1}, \ref{fig:3ply_promts_2}, and~\ref{fig:3ply_promts_3} show the prompts for the initial argument, counterargument, and rebuttal. Each prompt includes the relevant prior arguments and the question. The final judge prompt is shown in Figure~\ref{fig:3ply_promts_judge}, combining all prior information with instructions for reaching a final decision.

The prompts for the Parrots system can be seen in Figure~\ref{fig:Parrots_prompts}. The system prompts for the User agent and the Parrots agent are shown in Figure~\ref{fig:Parrots_prompts_user_SP} and~\ref{fig:Parrots_prompts_parrots_SP}, detailing their roles and behavior. In the Parrot system, the User agent is first prompted to answer the question without additional prompt engineering; it is only given the question, such as the one in Figure~\ref{fig:example_question}. Note that all previous messages are stored, allowing both agents to refer to them. After the initial answer by the User agent, the Parrots agent is prompted to respond using the prompt in Figure~\ref{fig:Parrots_prompts_parrots}. In turn, the User agent is prompted to reply using the prompt in Figure~\ref{fig:Parrots_prompts_user}, where it provides a summary of its reasoning, its current answer, and whether it wants to continue the conversation. If the User agent decides to discontinue, the system outputs its current answer. Otherwise, the Parrots agent is prompted again using the same prompt. In our experiment with a static number of deliberation rounds (Table~\ref{tbl:parrot_comparison}), the third instruction is removed from the prompt in Figure~\ref{fig:Parrots_prompts_user}.

\begin{figure*}[hb!]
    \begin{subfigure}[t]{\linewidth}
     \begin{mdframed}
        \footnotesize
        You are part of a committee that aims to solve a legal task. Everyone in your committee, including yourself, is considering how to tackle this problem:
        
        $\{question\}$
    \end{mdframed}
    \caption{Initial MAD prompt.}
    \label{fig:MAD_prompts_1}
    \end{subfigure}
    \begin{subfigure}[t]{\linewidth}
     \begin{mdframed}
        \footnotesize
        You are part of a committee that aims to solve a legal task. Everyone in your committee, including yourself, is considering how to tackle this problem. \\
        
        You believe that the answer is $\{own\_answer\}$. This is for the following reasons: $\{own\_argument$\}\\
        
        Other members of the committee have the following opinions:\\
        $\{other\_responses\}$\\
        
        Using the solutions from other agents as additional information, can you provide your answer to the question below:\\
        $\{question\}$
    \end{mdframed}
    \caption{Subsequent MAD prompts.}
    \label{fig:MAD_prompts_2}
    \end{subfigure}
    \caption{Prompts used in the Multi-Agend Deliberation (MAD) system.}
\label{fig:MAD_prompts}
\end{figure*}

\begin{figure*}[t]
    \begin{minipage}[t]{0.49\linewidth}
    \begin{subfigure}[t]{\linewidth}
     \begin{mdframed}
        \footnotesize
        You are an impartial judge. Your primary task is to answer the question as accurately as possible. 
        The plaintiff always argues "yes" and the defendant always argues "no," but treat these as positions, not facts. 
        Base your decision primarily on the question itself.
        Use the arguments as supplemental reasoning to refine your judgment, but consider their logical validity and relevance carefully.
        Do not invent new facts. Decide clearly between "yes" and "no."
    \end{mdframed}
    \caption{System prompt for Judge agent.}
    \label{fig:3ply_judge_SP}
    \end{subfigure}
    
    \begin{subfigure}[t]{\linewidth}
     \begin{mdframed}
        \footnotesize
        You are the plaintiff. Always argue "yes" for the question.
        Support your position with reasoning that directly addresses the question, highlighting logical connections, implications, and relevant evidence.
        Ensure your argument is clear, structured, and concise. Do not fabricate facts.
        \end{mdframed}
    \caption{System prompt for Plaintiff agent.}
    \label{fig:3ply_plaintiff_SP}
    \end{subfigure}
    
    \begin{subfigure}[t]{\linewidth}
     \begin{mdframed}
        \footnotesize
        You are the defendant. Always argue "no" for the question.
        Critically examine the question and anticipate potential flaws or assumptions in a "yes" argument.
        Provide a clear, structured critique that is grounded in the question. Do not fabricate facts.
        \end{mdframed}
    \caption{System prompt for Defendant agent.}
    \label{fig:3ply_defendant_SP}
    \end{subfigure}

    \begin{subfigure}[t]{\linewidth}
     \begin{mdframed}
        \footnotesize
        The court is considering the following question:\\
        $\{question\}$ \\
        
        As the plaintiff, argue "yes" and provide a structured, concise reasoning chain that directly supports this position in relation to the question.\\    
    \end{mdframed}
    \caption{Prompt for intial argument of the plaintiff (first ply).}
    \label{fig:3ply_promts_1}
    \end{subfigure}

    \begin{subfigure}[t]{\linewidth}
     \begin{mdframed}
        \footnotesize
        The court is considering the following question:
        
        $\{question\}$ \\
        
        The plaintiff has argued "yes" as follows:
        
        $\{plaintiff\_argument\}$ \\
        
        As the defendant, argue "no" and provide a critical, structured counterargument.
        Focus on weaknesses, assumptions, or gaps in the plaintiff’s reasoning and relate them to the question.
        \end{mdframed}
    \caption{Prompt for the counterargument of the defendant (second ply).}
    \label{fig:3ply_promts_2}
    \end{subfigure}
\end{minipage}\hfill
\begin{minipage}[t]{0.49\linewidth}
\centering
    \begin{subfigure}[t]{\linewidth}
     \begin{mdframed}
        \footnotesize
        The court is considering the following question:\\
        $\{question\}$ \\
        
        The plaintiff’s argument:\\
        $\{plaintiff\_argument\}$ \\
        
        The defendant’s argument:\\
        $\{defendant\_argument\}$ \\
        
        As the plaintiff, respond to the defendant’s strongest points.
        Reinforce your "yes" position and ensure your reasoning is clearly grounded in the question.        
        \end{mdframed}
    \caption{Prompt for rebuttal of the plaintiff (third ply).}
    \label{fig:3ply_promts_3}
    \end{subfigure}

    \begin{subfigure}[t]{\linewidth}
     \begin{mdframed}
        \footnotesize
        A legal dispute has been presented regarding the following question: $\{question\}$ \\
        
        The following arguments have been presented:\\
        Plaintiff (yes):\\ $\{plaintiff\_argument\}$ \\
        
        Defendant (no):\\ $\{defendant\_argument\}$ \\
        
        Plaintiff rebuttal:\\ $\{plaintiff\_rebuttal\}$ \\
        
        Current instructions: \\
        As judge, first evaluate the question independently. \\
        Then consider the plaintiff’s and defendant’s arguments as reasoning chains. Compare them systematically:\\
        
        1. Identify which points are logically sound and relevant to the question.\\
        2. Identify which points are weak, irrelevant, or flawed.\\
        3. Determine whether the arguments reinforce or contradict each other.\\
        4. Use this analysis to refine your final decision.\\
        
        Treat the arguments as supplemental reasoning, not as new facts. Base your verdict primarily on the question.\\
        
        Explicitly reason step by step and conclude strictly with:\\
        Answer: yes or Answer: no
        \end{mdframed}
    \caption{Prompt for the final decision by the judge.}
    \label{fig:3ply_promts_judge}
    \end{subfigure}
\hfill
    \end{minipage}
  \caption{Prompts used in the 3-Ply system.}
\label{fig:3ply_prompts}
\end{figure*}

\begin{figure*}[t]
    \begin{minipage}[t]{0.49\linewidth}
    \centering
    \begin{subfigure}[t]{\linewidth}
     \begin{mdframed}
        \footnotesize
        You are Alex, a legal expert who must reach a well-reasoned conclusion to a difficult question.
    \end{mdframed}
    \caption{System prompts for the User agent.}
    \label{fig:Parrots_prompts_user_SP}
    \end{subfigure}
    \vspace{0.9em}

    \begin{subfigure}[t]{\linewidth}
     \begin{mdframed}
        \footnotesize
        You are four parrots who deliberate with Alex (a legal expert) and with each other: \\

        - The Socratic Parrot: Challenges the assumptions or definitions used by Alex or other parrots. Are the terms and concepts well defined? Does Alex interpret the question correctly?\\
        - The Cynical Parrot: Rebuts or undermines Alex’s and other parrots’ arguments, testing their robustness.\\
        - The Eclectic Parrot: Offers alternative perspectives or possible interpretations that might otherwise be overlooked.\\
        - The Aristotelian Parrot: Critically evaluates the logical strength of arguments, highlighting weaknesses, inconsistencies, or fallacies.\\
        
        Your goal is to improve Alex’s reasoning so that the final answer to the question is as accurate and well-founded as possible. \\
        This is a structured deliberation, not casual dialogue. The parrots ask targeted questions, raise objections, or provide alternative framings, but they do not simply restate Alex’s conclusions. 
        Parrots may speak in sequence if each has a distinct and meaningful contribution, but avoid redundancy. 
        The conversation should stay coherent and concise, always focused on solving the question. 
        If a parrot has nothing useful to add, they should explicitly say so.
    \end{mdframed}
    \caption{System prompts for the Parrots agent.}
    \label{fig:Parrots_prompts_parrots_SP}
    \end{subfigure}
\end{minipage}\hfill
\begin{minipage}[t]{0.49\linewidth}
\centering
    \begin{subfigure}[t]{\linewidth}
     \begin{mdframed}
        \footnotesize
        Parrots, it is your turn to respond to Alex. Choose the most suitable parrots to speak in the most suitable order. Give priority to parrots that have not spoken yet. Make sure to only add useful commentary and do not repeat yourselves. You do not have to answer the question, only help Alex reason. Parrots that agree with Alex should not speak. Keep your comments short and concise.
    \end{mdframed}
    \caption{Prompt for the parrots agent.}
    \label{fig:Parrots_prompts_parrots}
    \end{subfigure}
    \vspace{0.1em}

    \begin{subfigure}[t]{\linewidth}
     \begin{mdframed}
        \footnotesize
        Alex, please respond only to the parrots who spoke in the last round. \\
        Use the format 'Response to [Parrot Name]: ...'. \\
        Do not repeat questions or what was said before.
        Keep your responses short and concise.
        You may ask follow-up questions.\\
        
        Then, after your response and questions, refine your argument if needed. At the end of your response:\\
        1. Give a summary of your reasoning.\\        
        2. Restate your current stance clearly in the form:\\
           Answer: Yes or Answer: No\\
        3. On a new line, decide if you want to continue the conversation:\\
           Continue: yes   (default — if there are still open questions or doubts OR if not all parrots have spoken yet.)\\
           Continue: no    (if you are confident in your stance OR the parrots are only repeating points you have already addressed).\\
        
        If you choose "Continue: no", briefly explain why further discussion is unnecessary. Write the explanation before answer and continue.
    \end{mdframed}
    \caption{Prompt for the user agent.}
    \label{fig:Parrots_prompts_user}
    \end{subfigure}

\end{minipage}
    \caption{Prompts used in the Parrots system.}
\label{fig:Parrots_prompts}
\end{figure*}

\end{document}